\documentclass[sigconf,nonacm,10pt]{acmart}
\setkeys{acmart.cls}{balance=false}
\usepackage{booktabs,array,graphicx,xcolor}
\usepackage{algorithmic}
\usepackage{algorithm}
\usepackage{xspace}
\usepackage{placeins}
\usepackage{xurl}
\usepackage{tikz}
\usetikzlibrary{arrows.meta,positioning,calc,fit,patterns,backgrounds}
\hypersetup{hidelinks}
\AtBeginEnvironment{thebibliography}{\raggedright}

\newcommand{\flowstep}[1]{{\fontfamily{cmr}\selectfont\textcircled{#1}}}

\makeatletter
\@ifclasslater{acmart}{2025/08/16}{}{%
  \def\@authornotemark{%
    \g@addto@macro\@currentauthors{%
      \advance\hfuzz by 5pt\relax\footnotemark\relax}}
  \def\@@authornotemark#1{%
    \g@addto@macro\@currentauthors{%
      \advance\hfuzz by 5pt\relax\footnotemark[#1]}}}
\makeatother

\begin{document}
\title[HBFresh]{\texorpdfstring{\fontsize{17.5pt}{20.5pt}\selectfont}{}HBFresh: Selective Refresh Management\texorpdfstring{\\}{ }for LLM Inference on High-Bandwidth Flash}

\author{Yuanwu Zhu}
\affiliation{%
  \institution{University of Science and Technology of China}
  \city{}
  \state{Anhui}
  \country{China}
}
\email{zzyuanyi@mail.ustc.edu.cn}

\author{YanPeng Hu}
\affiliation{%
  \institution{ShanghaiTech University}
  \city{Shanghai}
  \country{China}
}
\email{huyp@shanghaitech.edu.cn}

\author{Jingwei Sun\textsuperscript{*}}
\affiliation{%
  \institution{University of Science and Technology of China}
  \city{}
  \state{Anhui}
  \country{China}
}
\email{sunjw@ustc.edu.cn}

\author{Guangzhong Sun\textsuperscript{*}}
\affiliation{%
  \institution{University of Science and Technology of China}
  \city{}
  \state{Anhui}
  \country{China}
}
\email{gzsun@ustc.edu.cn}
\thanks{\textsuperscript{*}Corresponding author.}

\begin{abstract}
High-bandwidth flash (HBF) is an emerging NAND-based memory technology that expands memory capacity next to the GPU for LLM inference.
HBF holds model weights, which are written once and read many times, yet repeated reads accumulate read disturb and force refresh writes that consume program/erase endurance.
Whole-block refresh rewrites still-healthy pages.
Selective refresh rewrites fewer pages but leaves partially valid blocks that complicate later garbage collection (GC).
Our simulation of single-level-cell (SLC) read disturb with real model weights shows that a block serves a median of 4.2$\times$ the one-million-read interval of baseline whole-block refresh before its first page needs refresh.
Therefore, the refresh of healthy pages can be deferred.
We present HBFresh, which uses a Markov model of per-page placement and read-disturb history to forecast future free-space demand, and jointly decides where refreshed data is placed and when GC is triggered.
Simulations driven by LLM inference workloads show that HBFresh uses 18.75\%--60.6\% fewer page programs than the best static GC threshold at each occupancy and, at the lowest tested occupancy, lowers the worst-case inference latency from 73 ms to 17 ms.

\end{abstract}

\maketitle
\hypersetup{pdfauthor={Yuanwu Zhu, YanPeng Hu, Jingwei Sun, Guangzhong Sun}}
\pagestyle{plain}
\thispagestyle{plain}

\section{Introduction}\label{sec:introduction}
\begin{figure}[t]
\centering
\includegraphics[width=\columnwidth]{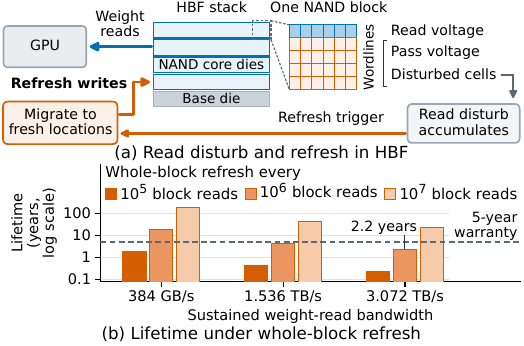}
\Description{A GPU reads weights from stacked NAND. The lower panel compares lifetime across bandwidths and refresh intervals.}
\caption{Refresh writes and SLC lifetime under sustained weight reads in HBF.}
\label{fig:motivation}
\end{figure}

High-bandwidth flash (\textbf{HBF}) offers read bandwidth comparable to high-bandwidth memory (\textbf{HBM}) at several times the capacity~\cite{ocp2026hbf,son2026hbf}. HBF designs store model weights, which are written once and read repeatedly, in HBF and keep frequently written data, such as the KV cache, in HBM~\cite{oliveira2026flint,ha2026h3,atassi2026hardware}. Such read-only placement is expected to avoid the write-bandwidth and endurance limits of NAND~\cite{saxena2027slipstream,oliveira2026flint}. However, reading unchanged weights still wears out HBF. Each page read disturbs the other pages of its block, and repeated reads cause read-disturb errors~\cite{cai2015readdisturb,wu2018characterizing}. The HBF specification of the Open Compute Project (\textbf{OCP}) therefore has host software rewrite a whole block, an operation called refresh~\cite{ocp2026hbf} (Fig.~\ref{fig:motivation}(a)). If a block is refreshed every $10^6$ reads at the maximum OCP-specified bandwidth, whole-block refresh exhausts the program/erase endurance of commodity single-level-cell (\textbf{SLC}) NAND~\cite{micron2016slc} in about 2.2 years (Fig.~\ref{fig:motivation}(b)). Thus, an HBF stack under whole-block refresh lasts less than half of the five-year warranty that HBF lifetime studies use~\cite{yang2026lachesis,baek2026hotcold}.

Whole-block refresh rewrites healthy pages, that is, pages that have not yet reached the refresh trigger (the bit-error count at which a page must be rewritten). Since the threshold voltages (the voltage at which a cell turns on) of commercial NAND are proprietary~\cite{cai2017errors}, we use a read-disturb simulator on real model weights and observe that pages of one block reach the refresh trigger at widely different times (Section~\ref{sec:opportunity}). Selective refresh rewrites only the pages at the refresh trigger into free pages of other blocks, so the refresh of healthy pages can be deferred. In SSDs, where read-disturb refresh is called read reclaim, recent policies also refresh only parts of a block~\cite{han2023ptdm,chun2025straw,lee2025dear}.

Selective refresh saves page programs (page writes) only if the healthy pages that selective refresh leaves in place are not migrated later. Valid pages hold the current copy of their data, and invalid pages hold an old copy. Because flash erases only whole blocks, garbage collection (\textbf{GC}) migrates the valid pages of a block to other blocks before erasing the block. Leaving healthy pages in place keeps their blocks, now partially valid, out of the free-block pool until GC migrates their remaining valid pages. Early GC migrates the healthy pages and cancels the page programs that selective refresh saves, while late GC can leave a later refresh without a free block. The SSD read-reclaim policies~\cite{han2023ptdm,chun2025straw,lee2025dear} do not adapt GC timing to partially valid blocks. The problem is therefore when to collect partially valid blocks and where to place refreshed pages, not how to refresh a page.

A static GC threshold selects a block for GC once its fraction of valid pages falls below the threshold. Occupancy is the fraction of device pages that hold weights. No static GC threshold is both cheap and safe across occupancies. The best static threshold, the cheapest fixed GC threshold that never runs out of free blocks, rises with occupancy (Section~\ref{sec:static}). An occupancy-tiered threshold is set by the current occupancy, and a low watermark on free blocks starts GC when free blocks fall to a fixed number. However, neither policy times GC well. First, at the same occupancy, the right threshold differs between block layouts (the numbers of valid, invalid and free pages in each block) and changes with the number of upcoming refreshes. Second, a watermark starts GC only when free blocks are scarce, so GC migrates many healthy pages at once.

Our key observation is that in HBF, the block layout, through its valid pages, sets how fast pages approach refresh. Classic analytic GC models~\cite{desnoyers2012analytic,vanhoudt2013meanfield} instead assume that writes invalidate pages uniformly at random and therefore miss this effect of the block layout. In a dense model, each decode step (the forward pass that generates the next token) reads every weight once~\cite{oliveira2026flint}, so all weight pages are read with the same frequency. Only valid pages are still read, so the read disturb of a page grows with the number of other valid pages in its block. Pages in partially valid blocks reach the refresh trigger more slowly, so these blocks stay partially valid longer. Hence, refresh demand and GC cost can be predicted accurately from the current block layout.

We present HBFresh, which refreshes only the pages at the refresh trigger and uses the predictable state evolution to decide when to collect blocks and where to place refreshed pages. In the backend, a Markov model of per-page placement and read-disturb history forecasts when and in which blocks future refreshes arise. Online, a refresh policy jointly decides where refreshed pages are placed and when GC runs. The Markov model estimates the probability of running out of free blocks for each candidate GC threshold, and the refresh policy adopts the cheapest candidate within a bound. To keep weight reads parallel after refresh, HBFresh also stripes refreshed pages across planes, the banks of a NAND die that serve reads in parallel.

The main contributions of this paper are as follows:
\begin{itemize}\setlength{\itemsep}{0pt}
\item We build the Markov model and a read-disturb simulator for SLC HBF, and use them to show that the right GC timing depends on the block layout and the upcoming refreshes.
\item We design HBFresh, which times GC with dynamic risk-bounded thresholds computed from the Markov model's forecasts, and stripes refreshed pages across planes.
\item In the read-disturb simulator, HBFresh reduces page programs by 18.75\%--60.6\% compared to the best static threshold at 25\%--75\% occupancy without running out of free blocks, and lowers the worst-case inference latency substantially, from 73~ms to 17~ms at 25\% occupancy. When pages are refreshed in logical-address order, striping them across planes restores the pre-refresh read bandwidth (Section~\ref{sec:striping}).
\end{itemize}

\section{Background and Motivation}\label{sec:background}
\subsection{HBF Organization and NAND Read Disturb}
An HBF stack places NAND core dies on a base die, the bottom die that manages the stack~\cite{ocp2026hbf}, and is co-packaged on a silicon interposer~\cite{oliveira2026flint} with the GPU that runs the model. Most HBF designs use SLC NAND, which stores one bit per cell~\cite{oliveira2026flint,son2026hbf,yang2026lachesis,yu2026characterizing,jonnalagadda2026splash}. During a read, the selected wordline (the row of cells that stores one page) receives the read reference voltage $V_{\mathrm{read}}$, while the unselected wordlines of the block receive a higher pass voltage $V_{\mathrm{pass}}$. As Fig.~\ref{fig:motivation}(a) shows, repeated exposure to $V_{\mathrm{pass}}$ shifts the threshold voltages of the unselected cells and causes read-disturb errors~\cite{cai2015readdisturb,spinelli2017reliability}. The disturbing reads of a page are reads to the other wordlines of its block, and a block read is a read to any of its pages.

\begin{figure}[t]
\centering
\includegraphics[width=\columnwidth]{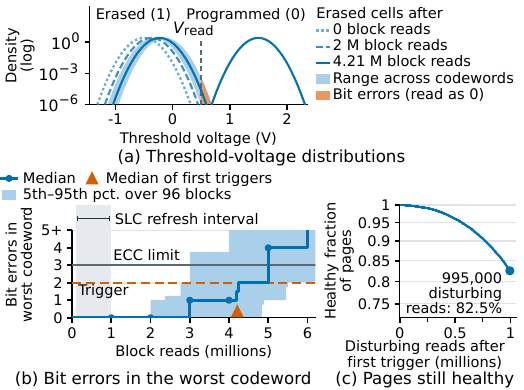}
\Description{Threshold-voltage distributions, bit-error counts versus block reads, and the healthy-page fraction after the refresh trigger.}
\caption{Read-disturb errors in SLC HBF blocks.}
\label{fig:headroom}
\end{figure}

Fig.~\ref{fig:headroom}(a) shows the threshold-voltage distributions of SLC cells. Erased cells, which store bit 1, lie below $V_{\mathrm{read}}$, and programmed cells, which store bit 0, lie above $V_{\mathrm{read}}$. Each disturbing read raises the threshold voltages of erased cells.

Erased cells above $V_{\mathrm{read}}$ are misread as 0 and become bit errors. A codeword is the part of a page that an error-correcting code (\textbf{ECC}) protects, together with its check bits. Because the threshold-voltage shift differs across codewords, the erased cells of different codewords in one block cross $V_{\mathrm{read}}$ at different times. The base die keeps a page-read counter per block and returns an ECC status with each read, either a correctable-error warning or an uncorrectable error~\cite{ocp2026hbf}. ECC corrects up to three bit errors per codeword. A page reaches the refresh trigger when its worst codeword has two bit errors, one below the ECC limit. We model the correctable-error warning as this trigger. HBF leaves refresh, data movement, and GC to host software~\cite{ocp2026hbf}.

\subsection{Refresh Writes under Sustained Weight Reads}
The refresh interval is the number of block reads after which host software refreshes a block~\cite{ocp2026hbf}. For SLC NAND, refresh intervals lie within $10^5$--$10^6$ block reads~\cite{oliveira2026flint,cai2015readdisturb}. At the maximum OCP-specified bandwidth of 3.072 TB/s~\cite{ocp2026hbf} and a refresh interval of $10^6$ block reads, every 256-page block of a 512 GiB stack whose blocks are read uniformly reaches $10^6$ block reads in about 11.7 minutes. Whole-block refresh therefore erases every block about 124 times a day. Fig.~\ref{fig:motivation}(b) shows the lifetime for the OCP bandwidths and refresh intervals from $10^5$ to $10^7$ block reads. The lifetime assumes uniform wear and $10^5$ program/erase cycles per block~\cite{micron2016slc,yu2026characterizing}.

\subsection{SSD Read Reclaim and HBF Refresh Policies}
\textbf{Refresh granularity, triggers, and placement.} Commercial SSDs mostly store two to four bits per cell. Flash Correct-and-Refresh periodically corrects pages and reprograms them in place or remaps their blocks to control retention errors~\cite{cai2012fcr}. PTDM migrates pages with read-count limits that depend on which bit of a multi-bit cell each page stores, for MLC and TLC flash~\cite{han2023ptdm}. However, SLC pages have no such page types. PTDM also spreads read-hot data across blocks in hot, warm, and cold classes~\cite{han2023ptdm}. PTDM's partial-block migration leaves blocks that cannot yet be erased, and PTDM collects them only with GC that runs when free blocks are scarce~\cite{han2023ptdm}. JIT-GC instead times background GC by estimating future SSD writes~\cite{hahn2015jitgc}. STRAW selects the wordlines to refresh with a stress model measured on 3D TLC chips~\cite{chun2025straw}, and DEAR refreshes wordlines from runtime error estimates~\cite{lee2025dear}. FineRR-ZNS adds host-side fine-grained read refresh to zoned SSDs and uses QLC flash by default~\cite{li2025finerrzns}. PhasedRR migrates hotter pages earlier~\cite{li2024phasedrr}, and SWEEP gathers read-hot and write-hot pages so that GC moves them~\cite{guo2026sweep}.

\textbf{Uniform weight reads.} Because all weight pages of a dense model are read with the same frequency~\cite{oliveira2026flint}, a dense model produces no read-hot pages in HBF for SSD hot/cold separation to exploit. If each token of a mixture-of-experts model picks 8 of 128 experts uniformly at random, a decode step with a batch of 64 tokens reads on average 98.4\% of the experts, so such a model also has almost no read-hot pages.

\textbf{HBF refresh and GC.} In concurrent work, Choi et al. also find that read reclaim limits HBF lifetime and that taller NAND stacks and higher bandwidth worsen this limit~\cite{choi2026hbfreadreclaim}. FLINT refreshes whole blocks~\cite{oliveira2026flint}. HBFlex and SPLASH stripe newly written KV pages across planes~\cite{zhong2026hbflex,jonnalagadda2026splash} but do not handle read-disturb refresh, and HBFlex starts GC at a fixed memory utilization~\cite{zhong2026hbflex}. None of these policies models how fast selective refresh consumes free blocks.

\subsection{Opportunity and Deferred Cost of Selective Refresh}\label{sec:opportunity}
We simulate 96 blocks that hold real Qwen3-32B BF16 weights, because read disturb shifts erased cells and the weight bits decide which cells stay erased. As Fig.~\ref{fig:headroom}(b) shows, no page reaches the refresh trigger within the SLC refresh interval of $10^5$--$10^6$ block reads. A block's first page reaches the refresh trigger after a median of 4.21 million block reads (5th--95th percentile 2.66--5.47 million). We then keep reading each block without refresh after its first trigger (Fig.~\ref{fig:headroom}(c)). After 995,000 more disturbing reads, \textbf{82.5\%} of the blocks' other pages are still healthy.

\begin{figure}[t]
\centering
\includegraphics[width=\columnwidth]{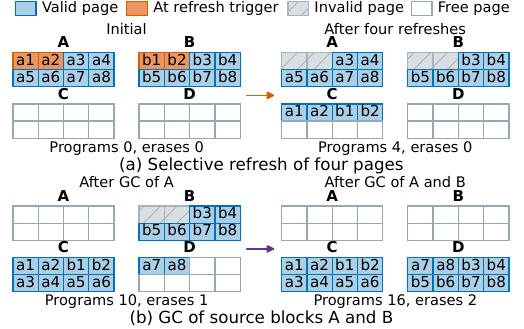}
\Description{Block grids show four page refreshes followed by garbage collection of two source blocks.}
\caption{Cost of refreshing four pages and later collecting their source blocks.}
\label{fig:sel-gc-example}
\end{figure}

In Fig.~\ref{fig:sel-gc-example}(a), refreshing four pages costs 4 page programs, but later collecting source blocks A and B in Fig.~\ref{fig:sel-gc-example}(b) adds 12 page programs and 2 erases. Thus, the two steps together cost as much as refreshing both whole blocks (16 page programs). In our simulator at 25\% occupancy, the static threshold of 0.75 likewise spends 5,120 of 8,448 page programs on GC. The healthy pages in Fig.~\ref{fig:headroom} and the GC cost in Fig.~\ref{fig:sel-gc-example} motivate a design that defers refreshing healthy pages and times GC by how fast selective refresh consumes free blocks.

\section{Read-Disturb Simulation and Markov Analysis}\label{sec:model}
\begin{figure}[t]
\centering
\includegraphics[width=\columnwidth]{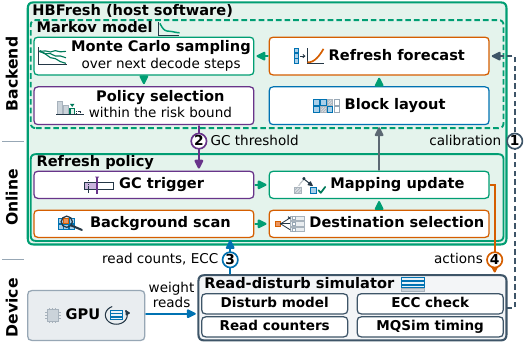}
\Description{The backend Markov model and online refresh policy exchange state and actions with the read-disturb simulator.}
\caption{Overview of HBFresh and the read-disturb simulator.}
\label{fig:design}
\end{figure}

\subsection{Overview}
To forecast the deferred GC cost of selective refresh over long runs and many policies, we pair the read-disturb simulator with a Markov model of per-page placement and read-disturb history. The read-disturb simulator stands in for the HBF device and directly measures the timing, page programs and erases of each policy.

The Markov model covers long runs and every occupancy and GC threshold, and returns the probability of running out of free blocks. Each transition of the Markov model is one refresh event: a page reaches the refresh trigger and is refreshed, followed by any GC that the refresh causes.

HBFresh runs on the host with a backend Markov model and an online refresh policy (Fig.~\ref{fig:design}). The simulator supplies calibration data to the Markov model (\flowstep{1}), which selects a GC threshold for the refresh policy (\flowstep{2}). In place of the base die, the simulator reports read counts and ECC status (\flowstep{3}) and runs the policy's actions (\flowstep{4}).

\subsection{Read-Disturb Simulator}
The read-disturb simulator maps the bits of a model checkpoint to NAND pages. During programming, each cell's threshold voltage is sampled from the Gaussian distribution of the erased or programmed state (Fig.~\ref{fig:headroom}(a)). The simulator adds block-level and wordline-level variation~\cite{di2016exploiting,nie2020layer} and a fraction of weak blocks, whose read-disturb shift is 2.5$\times$ that of other blocks. Each page read shifts the threshold voltages of the unread pages in the same block.

The read-disturb count of a page is the number of disturbing reads that the page has received since it was programmed at its current location. Therefore, a migrated copy starts from zero, whereas healthy pages left in place keep their counts.

\begin{table}[t]
\centering
\caption{Notation}
\label{tab:notation}
\footnotesize
\setlength{\tabcolsep}{3pt}
\begin{tabular}{@{}>{\raggedright\arraybackslash}p{0.14\columnwidth}>{\raggedright\arraybackslash}p{0.83\columnwidth}@{}}
\toprule
$h_i$, $P_{\mathrm{blk}}$ & Valid pages in block $i$; pages per block.\\
$u_i$ & Programmed pages in block $i$.\\
$\eta$ & Refresh rate of a page per disturbing read (calibrated).\\
$\lambda_i$, $c$, $r_s$ & Refreshes of block $i$ per decode step, $c\,h_i(h_i-1)$, with $c=\eta r_s$ and $r_s$ reads of a valid page per decode step (1 if dense).\\
$\theta$ & GC threshold on the fraction of valid pages of a closed block.\\
$\Pi$, $\pi$, $\pi^\star$ & Candidate values of $\theta$, including no GC; one candidate; the adopted one.\\
$Q$, $\epsilon$ & Horizon in decode steps; bound on the probability of running out of free blocks within $Q$ decode steps.\\
\bottomrule
\end{tabular}
\end{table}

\subsection{Markov State: Layout and Read-Disturb History}
\begin{figure*}[t]
\centering
\includegraphics[width=\textwidth]{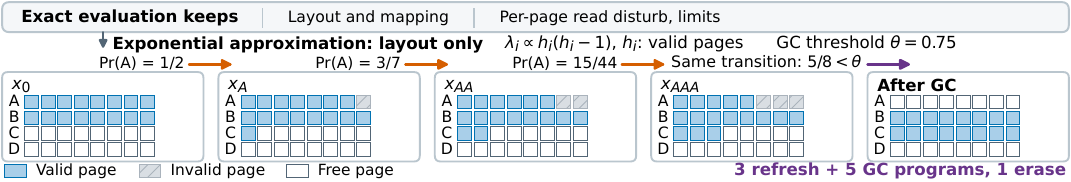}
\Description{State diagrams trace three refresh transitions and garbage collection of a source block.}
\caption{Markov state and a worked example of refresh and GC.}
\label{fig:capacity}
\end{figure*}

The Markov state keeps only what determines the next transition. Evaluated exactly, the Markov model keeps the layout, the page mapping, and the read-disturb count and limit of each page. The read-disturb limit of a page is the count at which the page reaches the refresh trigger.

For each block $i$ of $P_{\mathrm{blk}}$ pages, the layout records the number of valid pages $h_i$ and the number of programmed pages $u_i$ (Table~\ref{tab:notation}) and thus distinguishes the $h_i$ valid pages, the $u_i-h_i$ invalid pages and the $P_{\mathrm{blk}}-u_i$ free pages of block $i$ (Fig.~\ref{fig:capacity}). The layout also records whether block $i$ is free, open (still accepting writes), closed, or reserved (Section~\ref{sec:reserve}). Invalid pages become free only when the block is erased. Since only an open block accepts writes into its free pages, the Markov model can tell whether the next refresh fits into an open block or a new block must be opened.

A refresh moves one page from source block $i$ to a free page of destination block $j$: $h_i$ drops by one while $u_i$ stays the same, and $h_j$ and $u_j$ grow by one. Consequently, block $i$ gains no free page until GC migrates its remaining valid pages and returns it to the free-block pool.

\subsection{Transition Probabilities and a Worked Example}
HBFresh evaluates the Markov model under an exponential approximation. In this approximation, every valid page is read $r_s$ times per decode step and the reads left before refresh are independent exponential variables with rate $\eta$ per disturbing read. Refreshes then arise in block $i$ at rate $\lambda_i$ (Eq.~\eqref{eq:rates}), because each of the $h_i$ valid pages of block $i$ receives disturbing reads from the other $h_i-1$.
\begin{equation}
\lambda_i=\eta r_s h_i(h_i-1)=c h_i(h_i-1),
\quad c=\eta r_s.
\label{eq:rates}
\end{equation}
Here $c=\eta r_s$ is the refresh rate that one valid neighbor adds to a page, so $\lambda_i$ counts refreshes per decode step. For example, by Eq.~\eqref{eq:rates}, a full block of eight pages has rate $56c$, whereas a block with four valid pages has rate $12c$, so refreshes arise more slowly in a partially valid block.

The layout alone forms the whole state, because every valid page is read with the same frequency and exponential variables are memoryless, so neither the mapping nor the reads a page has already received change the distribution of the reads left before its refresh. The time to the next refresh event is exponential with rate $\sum_j\lambda_j$ per decode step. Summing these exponential times gives the number of refresh events that fall within the horizon of $Q$ decode steps (Table~\ref{tab:notation}). The per-page read-disturb history enters through the rate $\eta$, which we calibrate against the simulator (\flowstep{1} in Fig.~\ref{fig:design}). HBFresh re-plans from the current layout after each refresh event.

With these rates, the next refresh comes from block $i$ with probability $\lambda_i/\sum_j\lambda_j$. Because the refresh policy fixes where each refresh goes and which blocks GC collects, the probabilities $\lambda_i/\sum_j\lambda_j$ define a Markov chain over layouts.

The worked example in Fig.~\ref{fig:capacity} starts from blocks A--D of Fig.~\ref{fig:sel-gc-example}(a). Blocks A and B are closed and full, with 16 logical pages in total. Blocks C and D are free. A closed block is collected once its fraction of valid pages falls below the GC threshold $\theta=0.75$. Each arrow label is the probability that the next refresh comes from that block. In the initial layout $x_0$, A and B each have rate $56c$ (Eq.~\eqref{eq:rates}), so each supplies the first refresh with probability 1/2. After two refreshes from A, a third refresh from A leaves five valid pages in A and three in C ($x_{AAA}$). Since $5/8<\theta$, the same transition migrates the five valid pages of A to C and returns A to the free-block pool. The sequence costs three refresh programs, five GC programs, and one erase.

\subsection{Free-Block Reserve and Policy Selection}\label{sec:reserve}
HBFresh keeps a free-block reserve of 5\% of the blocks, rounded up to whole blocks, to leave room for later writes, such as the KV cache in designs that place it in HBF~\cite{yang2026lachesis}.

Among the candidate GC thresholds (Section~\ref{sec:gctiming}) whose probability of running out of free blocks stays within a bound, Eq.~\eqref{eq:risk} selects the one with the fewest expected page programs.
\begin{equation}
\begin{aligned}
\mathrm{Risk}_Q(\pi\mid\widehat{x})
 &=\Pr(q_{\mathrm{fail}}\le Q\mid\widehat{x},\pi),\\
\pi^\star&\in\arg\min_{\pi\in\Pi}
\mathbb{E}[N_{\mathrm{prog}}(Q)\mid\widehat{x},\pi]\\
&\qquad\text{s.t.}\ \mathrm{Risk}_Q(\pi\mid\widehat{x})\le\epsilon.
\end{aligned}
\label{eq:risk}
\end{equation}
In Eq.~\eqref{eq:risk}, $\widehat{x}$ is HBFresh's estimate of the current Markov state. Table~\ref{tab:notation} defines $\Pi$, $\pi$, $Q$ and $\epsilon$. The device runs out of free blocks when a migration finds no free page in any block it may write. The variable $q_{\mathrm{fail}}$ counts decode steps from the decision until the device first runs out of free blocks and is infinite if this never happens, and $N_{\mathrm{prog}}(Q)$ counts the refresh and GC programs issued within the $Q$ decode steps after the decision.

\subsection{Static GC Thresholds Across Occupancies}\label{sec:static}
\begin{figure}[t]
\centering
\includegraphics[width=\columnwidth]{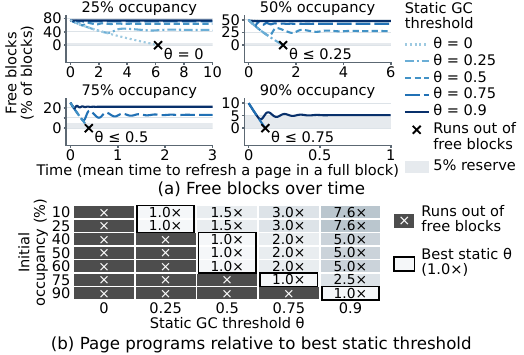}
\Description{Free blocks over time for static thresholds and a grid of page-program ratios marked where free blocks run out.}
\caption{Free blocks and page programs at static GC thresholds.}
\label{fig:static-capacity}
\end{figure}

The best static threshold rises with occupancy. We evaluate the Markov model under the exponential approximation on a 512 GiB device across occupancies and GC thresholds. Fig.~\ref{fig:static-capacity}(a) shows free blocks over time at four occupancies, with time in units of $1/((P_{\mathrm{blk}}-1)c)$, the mean time until a page in a full block reaches the refresh trigger (Eq.~\eqref{eq:rates}). For example, at 50\% occupancy, $\theta\le0.25$ runs out of free blocks before any block is collected, whereas with $\theta=0.5$ the free-block fraction falls to a minimum of 18.5\%. With the higher threshold $\theta=0.75$, more free blocks remain, but each refresh costs twice as many page programs as with $\theta=0.5$. Across occupancies (Fig.~\ref{fig:static-capacity}(b)), the best static threshold is 0.25 at 10\%--25\%, 0.5 at 40\%--60\%, 0.75 at 75\%, and 0.9 at 90\% occupancy. Without GC ($\theta=0$), free blocks run out at all evaluated occupancies. A higher threshold keeps more free blocks but costs more page programs per refresh.

\section{HBFresh Refresh Management}\label{sec:design}
\subsection{Overview}
Since no static GC threshold is both cheap and safe across occupancies (Section~\ref{sec:static}), we design HBFresh, which runs on the host (Fig.~\ref{fig:design}), to derive GC thresholds from the current layout. During inference, the refresh policy refreshes the pages that reach the refresh trigger, and healthy pages continue to serve reads in place. HBFresh needs two inputs from the base die: the per-block read counts and the ECC status of each page read. A background scan reads pages to check their ECC status.

\subsection{Risk-Bounded GC Timing}\label{sec:gctiming}
The candidate GC thresholds include no GC plus the thresholds that the Markov model derives from the current layout. Each derived threshold is a valid-page fraction for which collecting the closed blocks below it restores the free-block reserve. In the backend, the Markov model evaluates each candidate by Monte Carlo sampling of the Markov chain over the next $Q$ decode steps (Fig.~\ref{fig:design}). The refresh policy adopts the candidate that Eq.~\eqref{eq:risk} selects. Each static threshold also runs watermark GC, which starts when free blocks fall to the free-block reserve, and HBFresh uses it only if no candidate qualifies. A policy query runs in the backend, off the inference path. Until a new threshold is adopted, HBFresh keeps the previous one.

\subsection{Destination Selection Across Planes}
HBF reaches its bandwidth by reading many planes in parallel, and each block lies in one plane. Whole-block refresh copies a block into one free block~\cite{ocp2026hbf,oliveira2026flint}, so pages that were read in parallel stay on different planes. In contrast, selective refresh gathers pages from many blocks, and appending them to one open block would concentrate their later reads on one plane. We therefore send each refreshed page to one of up to $K$ open destination blocks on different dies. The destination is chosen by the page's original block index plus its page offset, modulo $K$, so the gathered pages spread over $K$ dies and hence over $K$ planes. SSD superblocks similarly link blocks across planes~\cite{wang2019was}.

\subsection{Data Migration and Mapping Update}
Each migration programs ECC-corrected data into a free page, or uses data from a backup copy of the weights on an SSD if the error is uncorrectable~\cite{ocp2026hbf}. HBFresh then reads the page back to verify it. If the mapping entry still points to the old page, HBFresh atomically redirects the entry to the new page and marks the old page invalid. To collect a block, HBFresh migrates its remaining valid pages the same way, and the next write to page 0 of the block triggers an automatic block erase in HBF~\cite{ocp2026hbf}.

\section{Evaluation}\label{sec:evaluation}

\subsection{Experimental Setup and Metrics}
\begin{table}[t]
\centering
\caption{Simulation setup}
\label{tab:setup}
\footnotesize
\setlength{\tabcolsep}{3pt}
\begin{tabular}{@{}>{\raggedright\arraybackslash}p{0.37\columnwidth}>{\raggedright\arraybackslash}p{0.60\columnwidth}@{}}
\toprule
\multicolumn{2}{@{}l}{\textit{Media model}} \\
Erased / programmed mean & $-0.5$~V / 1.5~V, std. dev. 0.15~V (Gaussian) \\
Read reference voltage & 0.5~V \\
Block / wordline std. dev. & 0.03~V / 0.02~V, weak blocks 5\% \\
Shift per $10^6$ reads & 0.65--1.35$\times$ nominal 0.065~V (per codeword) \\
\midrule
\multicolumn{2}{@{}l}{\textit{HBF device and policy runs}} \\
Device & 8 GiB, 8,192 blocks $\times$ 256 pages, 16 dies $\times$ 32 planes, 1.536 TB/s \\
Injected errors & one every 512 decode steps, 4 in total, each into 0.15\%--0.2\% of the weight pages \\
Mixture-of-experts run & 15 MiB mixture-of-experts model (8 experts, top-2), 16 dies $\times$ 16 planes, 3.072 TB/s, 192,000 tokens \\
Markov model & fed with refreshes from simulator runs not used for calibration \\
\midrule
\multicolumn{2}{@{}l}{\textit{Striping and layouts}} \\
Striping & Synthetic dense model on the same 8~GiB device, 8,192 pages rewritten in logical-address order or 24,576 in random order, $K$ open blocks \\
Layouts & 64 blocks, 5 free at start, 4-block reserve, 4 half-valid or 16 blocks with 224 valid pages, other source blocks full, 256 or 768 refreshes \\
\bottomrule
\end{tabular}
\end{table}

\begin{figure}[t]
\centering
\includegraphics[width=\columnwidth]{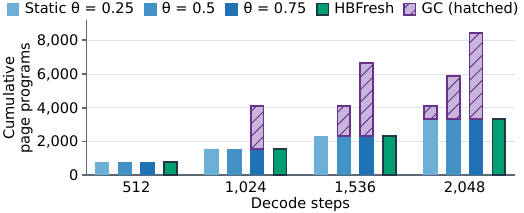}
\Description{Page programs over successive decode steps. Hatching distinguishes programs due to garbage collection.}
\caption{Cumulative refresh and GC programs per policy at 25\% occupancy.}
\label{fig:programs-round}
\end{figure}

\textbf{Platform.} We measure latency, bandwidth, page programs and migrations (Figs.~\ref{fig:programs-round}--\ref{fig:bandwidth-k}) with the read-disturb simulator (Table~\ref{tab:setup}), which uses MQSim, a multi-queue SSD simulator~\cite{tavakkol2018mqsim}, for timing. We check whether free blocks run out over long runs with the Markov model of Section~\ref{sec:model}. Since absolute threshold voltages are proprietary~\cite{cai2017errors}, the voltages in Table~\ref{tab:setup} form an illustrative scale. With these voltages, no page reaches the refresh trigger before $10^6$ block reads, consistent with SLC flash, where read-disturb errors are expected only after that many reads to a block~\cite{cai2015readdisturb}. Halving or doubling the read-disturb shift moves the median first-trigger read count in Fig.~\ref{fig:headroom}(b) from 4.21 million to 8.42 or 2.11 million block reads. Changing the threshold-voltage spread by 10\%, or the fraction of weak blocks from 5\% to 0\% or 10\%, keeps the median between 3.55 and 4.91 million, all above $10^6$.

\textbf{Policy runs.} Each decode step of a dense model reads every weight page once. Therefore, an 8 GiB device sees the same number of reads per block in each decode step as a 512 GiB device. Pages reach the refresh trigger through read disturb in the media model or through errors we inject at a fixed interval (Table~\ref{tab:setup}), since read disturb alone brings no page to the refresh trigger within the simulated reads (Section~\ref{sec:programs}). Since each injection places errors into a new pseudo-random set of weight pages that is the same for every policy, all policies see identical refreshes, and the comparison isolates GC timing.

\textbf{Baselines.} Whole-block refresh rewrites a block every $10^6$ block reads. Static-threshold GC refreshes pages at the refresh trigger, like DEAR~\cite{lee2025dear}, and stripes them over $K$ open destination blocks as HBFresh does, but collects blocks below a static GC threshold $\theta$. For the layout comparison, we use watermark GC and an occupancy-tiered threshold whose tiers are set on a different block layout. All policies share the workload, initial mapping, and capacity.

\textbf{Metrics.} We report page programs by cause (Fig.~\ref{fig:programs-round}). Refresh programs rewrite pages at the refresh trigger, and GC programs move healthy pages out of collected blocks. Initial loading is excluded. Erases include the first erase of each destination block. Figs.~\ref{fig:sel-gc-example} and~\ref{fig:capacity} count only erases of collected blocks. We report the inference latency per decode step.

\subsection{Page Programs and Inference Cost}\label{sec:programs}
Under read-disturb errors alone, HBFresh issues no page program and no erase for a 1 GiB synthetic dense model on the device of Table~\ref{tab:setup}, because no page reaches the refresh trigger. HBFresh costs only the reads of the background scan, a negligible 0.026\% of the inference reads. With the complete \mbox{Qwen3-1.7B} model on a 512 GiB device, only 0.25\% of the model weights reach the refresh trigger after 1.8$\times$ the whole-block refresh interval. In a mixture-of-experts run where pages do reach the refresh trigger (Table~\ref{tab:setup}), selective refresh with a round-robin background scan and watermark GC issues only 3,145--3,899 page programs and 16--18 erases. Whole-block refresh issues the same 28,416 page programs and 118 erases in every run, because its cost follows the read count alone.

Deferring GC lets HBFresh write fewer pages and avoid a concentrated GC burst. Each static threshold starts GC as its source blocks lose valid pages, whereas HBFresh defers GC. At 25\% occupancy, HBFresh uses \textbf{18.75\%} fewer page programs than the best static threshold of 0.25 (Fig.~\ref{fig:programs-round}). HBFresh also lowers the worst-case inference latency from 73.12 ms for the best static threshold to 16.98 ms. The best static threshold concentrates all its GC in one burst, and this concentrated GC raises its worst-case inference latency.

\begin{figure}[t]
\centering
\includegraphics[width=\columnwidth]{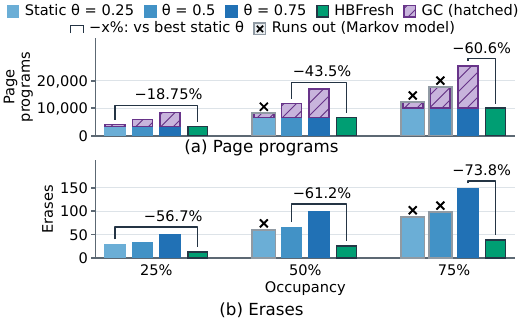}
\Description{Bars show page programs and erases for four policies at three occupancies. Crosses mark exhausted free blocks.}
\caption{Page programs and erases per policy at 25\%--75\% occupancy.}
\label{fig:programs-occupancy}
\end{figure}

HBFresh's margin grows with occupancy because the best static threshold rises with occupancy, and a higher threshold collects blocks that still hold more healthy pages. At the higher occupancies of 50\% and 75\%, HBFresh uses 43.5\% and 60.6\% fewer page programs than the best static thresholds of 0.5 and 0.75, respectively (Fig.~\ref{fig:programs-occupancy}(a)). Across 25\%--75\% occupancy, HBFresh uses 56.7\%--73.8\% fewer erases (Fig.~\ref{fig:programs-occupancy}(b)), and each erase consumes one of the program/erase cycles that bound HBF lifetime. Each static threshold marked $\times$ in Fig.~\ref{fig:programs-occupancy} runs out of free blocks in the Markov model, so the next refresh stalls (Fig.~\ref{fig:static-capacity}(b)), whereas HBFresh never runs out of free blocks in these runs.

\subsection{GC Timing Across Layouts and Destination Striping}\label{sec:striping}

\begin{figure}[t]
\centering
\includegraphics[width=\columnwidth]{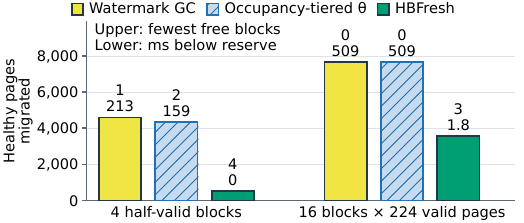}
\Description{Bars compare the healthy pages migrated by the three policies under two layouts.}
\caption{Healthy pages migrated by each GC policy after 768 refreshes.}
\label{fig:layouts}
\end{figure}

At the same occupancy, the Markov model times GC according to the layout and the upcoming refreshes. In the Markov model, on a 256-block device with 16 free blocks and a 13-block reserve, we compare two layouts: four half-valid blocks with 128 valid pages each, or 16 blocks with 224 valid pages each. With 256 upcoming refreshes, deferring GC keeps at least 15 free blocks in both layouts, and the Markov model chooses no GC. With 1,024 upcoming refreshes, deferring GC leaves only 12 free blocks in both layouts, below the 13-block reserve. The Markov model then chooses to collect the four half-valid blocks and the 16 blocks with 224 valid pages, respectively, which keeps at least 14 free blocks. In contrast, an occupancy-tiered threshold gives the same answer in all four cases.

After 768 refreshes on a simulated 64-block device with either layout (Table~\ref{tab:setup}), the collection that the Markov model selects (HBFresh in Fig.~\ref{fig:layouts}) migrates 512--3,584 healthy pages, whereas watermark GC and the occupancy-tiered threshold migrate 4,352--7,680. HBFresh returns to the 4-block reserve within 1.8 ms, whereas the two baselines fall to 0--2 free blocks and stay below the reserve for 159--509 ms. With 256 refreshes, free blocks stay at the 4-block reserve without GC, whereas applying the collection selected for 768 refreshes would migrate 3,584 healthy pages in the layout of 16 blocks with 224 valid pages.

\begin{figure}[t]
\centering
\includegraphics[width=\columnwidth]{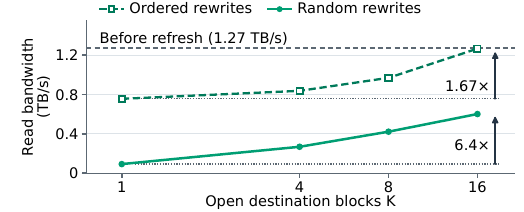}
\Description{Curves compare bandwidth against the number of open destination blocks for ordered and random rewrites.}
\caption{Read bandwidth after refresh versus open destination blocks.}
\label{fig:bandwidth-k}
\end{figure}

Striping refreshed pages across open destination blocks keeps weight reads parallel after refresh (Fig.~\ref{fig:bandwidth-k}). We refresh pages either in logical-address order (ordered rewrites) or in random order (random rewrites). With $K=16$, striping restores the pre-refresh bandwidth of 1.27 TB/s after ordered rewrites, 1.67$\times$ the bandwidth at $K=1$. After random rewrites, $K=16$ gives 6.4$\times$ the bandwidth at $K=1$.

\subsection{Model Accuracy and Overhead}
When we feed the refreshes that the simulator reports (Table~\ref{tab:setup}) to the Markov model in its exponential approximation, the model picks the policy with the fewest page programs in the simulator at every occupancy. Moreover, evaluated exactly on a dense model, the Markov model matches the simulator's refresh count within 7.9\%.

HBFresh runs the Markov model's forecast asynchronously in the background. In a paired simulation of the same workload, the inference latency stays the same with and without the forecast, and inference does not need to wait for a forecast result.

\section{Conclusion}\label{sec:conclusion}
We presented HBFresh, which refreshes only pages at the refresh trigger and times GC with a Markov model of per-page placement and read-disturb history. In the uniform weight-read setting of HBF, the block layout changes caused by selective refresh further change the subsequent read disturb, refresh demand, and GC cost. HBFresh exploits this predictable state evolution to decide when to collect and where to place pages, thereby controlling both write overhead and the risk of space exhaustion. HBFresh uses 18.75\%--60.6\% fewer page programs than the best static GC threshold at each occupancy and, at the lowest tested occupancy, lowers the worst-case inference latency from 73 ms to 17 ms.

\bibliographystyle{ACM-Reference-Format}
\bibliography{references}
\end{document}